\documentclass[letterpaper,10pt,conference]{ieeeconf}
\IEEEoverridecommandlockouts
\usepackage{booktabs}
\usepackage{multirow}
\usepackage{tabularx}
\usepackage{graphicx}
\usepackage{xcolor}
\usepackage{amsmath}
\usepackage{amssymb} 
\usepackage{amsfonts}
\usepackage{cite}
\makeatletter
\let\NAT@parse\undefined
\makeatother
\usepackage[hidelinks]{hyperref}
\usepackage{comment}
\graphicspath{{figures/}}
\usepackage{etoolbox}
\makeatletter
\patchcmd{\@makecaption}
  {{\footnotesize #1}\\{\footnotesize\scshape #2}}
  {{\footnotesize #1: {\scshape #2}}}
  {}{}
\makeatother

\title{\LARGE\bf GUIDE: Goal-Initialized Directional Understanding for End-to-End Legged Navigation}

\author{
Liang Wang$^{1,2,*}$, Jin Jin$^{3,*}$,
Kanzhong Yao$^{2}$, Yibin Wu$^{3,4}$,
Fangqiang Ding$^{5}$,\\
Jin Wang$^{3}$, Jun Wu$^{1}$,
Zhe Sun$^{2,\dagger}$, Qiuguo Zhu$^{1,\dagger}$
\thanks{$^{*}$Equal contribution.
$^{\dagger}$Corresponding authors.}
\thanks{
$^{1}$Institute of Cyber-Systems and Control, Zhejiang University;
$^{2}$Institute of Artificial Intelligence (TeleAI), China Telecom;
$^{3}$University of Oxford;
$^{4}$Center for Robotics, University of Bonn;
$^{5}$The Hong Kong University of Science and Technology (Guangzhou).
}
}

\begin{document}
\maketitle
\thispagestyle{empty}
\pagestyle{empty}

\begin{abstract}

    End-to-end reinforcement learning (RL) has shown strong potential for legged robot navigation, yet existing approaches commonly rely on continuously updated robot-to-goal states from external state estimation modules, leaving part of the navigation problem outside the learned policy. In this work, we seek to push the limits of end-to-end sim-to-real RL navigation by investigating whether a legged robot can internally maintain the spatial context required for long-horizon navigation. To this end, we study goal-initialized navigation, where the goal is provided only once at the beginning of an episode, with no subsequent external relative-goal updates. We present GUIDE, an end-to-end RL framework that jointly learns navigation and internal directional awareness from onboard observations. GUIDE leverages multi-frequency proprioceptive history to capture egomotion and auxiliary spatial-anchor prediction to maintain task-relevant spatial states, while temporal depth observations provide local environmental geometry. The navigation policy is trained entirely in simulation and transferred zero-shot to the real world. Across cluttered environments and structured mazes, GUIDE reliably avoids obstacles, escapes dead ends, and reaches distant goals using only onboard sensing. These results demonstrate that robust sim-to-real legged navigation can be achieved without continuously providing external robot-to-goal estimation, opening a promising direction for future research on more self-contained end-to-end navigation. Video demonstrations are available at: \url{https://guide-navigation.github.io/}.
    
\end{abstract}

\section{Introduction}

Legged robots possess inherent agility to traverse complex terrains and navigate in confined environments~\cite{hoeller2024anymal, cheng2024extreme, wu2025iros, ToFNavRAL26}. 
Such environments often contain narrow passages, occlusions, unexplored regions, and dead ends, where onboard sensing provides only partial observations of the surroundings. 
Robust navigation therefore requires not only reactive obstacle avoidance, but also the ability to maintain spatial context over extended horizons and avoid myopic decisions that may lead to locally attractive but globally unproductive states~\cite{yang2025spatially, jeong2026hipan}.

End-to-end learning provides a promising approach to this problem by directly mapping onboard observations to navigation actions~\cite{tai2017virtual, Wijmans2019DDPPOLN, shi2019end, huang2023goal}. 
Recent reinforcement learning (RL) methods have demonstrated increasingly capable obstacle avoidance and long-range navigation behaviors~\cite{yang2025spatially, lee2024learning, yuan2025reasan}. 
However, despite learning the navigation policy end-to-end, these approaches commonly rely on continuously updated robot-to-goal states provided by external localization or state estimation modules. 
Consequently, an important part of the navigation problem---maintaining the robot's evolving spatial relation to the goal---remains outside the learned policy. 
This raises a natural question when pushing the limits of end-to-end sim-to-real RL navigation: \emph{can a legged robot navigate over long horizons without continuous external robot-to-goal updates by maintaining the required spatial context internally?}

\begin{figure}[!t]
    \centering
    \includegraphics[width=\linewidth]{figures/hero_compressed.pdf}
    \vspace{-1em}
    \caption{GUIDE cultivates internal egomotion and directional awareness for end-to-end quadruped navigation, allowing the policy to traverse complex maze-like terrains and escape dead ends with the target location given only at initialization. The central simulation inset illustrates joint learning of navigation and spatial awareness. The top and bottom panels show wide and side views of a real-world deployment, where the gradient-colored curve traces the robot's trajectory through multiple dead ends to the goal.
}
    
    \label{fig:teaser}
\vspace{-2em}
\end{figure}

To investigate this question, we study \textbf{goal-initialized navigation}, where the goal is provided only once at the beginning of an episode and no subsequent external relative-goal updates are available to the policy. 
This setting makes long-horizon navigation particularly challenging under partial observability. 
The robot must continuously reason about how its own motion changes its relation to the initial goal, while simultaneously using local observations to avoid obstacles, recognize unproductive routes, and escape dead ends. 
Importantly, legged robots naturally provide rich proprioceptive histories through their joint encoders and inertial sensors, which contain the motion information of the robot~\cite{wasserman2025legolas, 11375910}. 
This motivates us to investigate whether task-relevant egomotion and directional context can be learned directly from proprioception along with the navigation policy rather than continuously supplied by an external estimator.

To this end, we propose GUIDE, an end-to-end RL framework that jointly learns navigation and internal directional awareness from onboard observations. 
GUIDE processes multi-frequency proprioceptive history to capture both long-term motion context and fine-grained egomotion, while temporal depth observations encode the geometry information of the environment. 
During training, an auxiliary spatial-anchor predictor supervises the proprioceptive representation to estimate task-relevant spatial states, including the relative goal location, relative spawn position, and body velocity in the ego-centric frame. 
These learned representations are integrated with depth observations to support long-horizon navigation without external streaming relative-goal inputs. 
The navigation policy is trained entirely in simulation and transferred zero-shot to a real quadruped robot. 
Through extensive simulation and real-world experiments, we show that GUIDE can navigate dense clutter, escape structural dead ends, and traverse challenging maze environments using only onboard observations after goal initialization, as shown in Fig.~\ref{fig:teaser}. The main contributions of this work are summarized as follows:
\begin{itemize}
    \setlength{\itemsep}{2pt}
    \setlength{\parsep}{0pt}
    \setlength{\parskip}{0pt}

    \item We pursue the limit of sim-to-real RL based end-to-end navigation. We study goal-initialized navigation for quadruped robots, where the goal is provided only at initialization, and no external streaming relative-goal updates are available during navigation.

    \item We propose GUIDE, an end-to-end RL framework that combines temporal depth perception with multi-frequency proprioceptive history and auxiliary spatial-anchor prediction, enabling the policy to learn task-relevant egomotion and directional context internally.

    \item We validate GUIDE extensively in simulation and demonstrate zero-shot transfer to real-world environments, showing robust goal reaching and dead-end escaping using only onboard sensing.
\end{itemize}

\section{Related Work}
\label{sec:citations}

\subsection{Learning-Based Local Navigation}

Recent works in imitation learning leverage expert demonstrations or large-scale offline datasets to map observations directly to control commands~\cite{sadat2020perceive, Pfeiffer2016FromPT, shah2023gnm, Shah2023ViNTAF}, which allow robots to inherit expert-level decision-making but are typically confined to the specific data distributions. Imperative learning paradigms~\cite{yang2023iplanner, roth2024viplanner} address local navigation by optimizing differentiable objectives to generate collision-free and semantically-informed paths from onboard perception. Complementary to these, reinforcement learning (RL) is widely adopted to discover agile obstacle avoidance and terrain traversal strategies through large-scale interaction with simulated environments~\cite{hoeller2021learning, zhang2024resilient, he2024agile, roth2025fdm,  chen2026seanavefficientpolicylearning}. Despite their proficiency in local maneuvers, these methods may still make myopic decisions and become trapped in dead ends under partial observability~\cite{long2017deep, long2018towards, ding2018hierarchical}.

\subsection{Long-horizon Mapless Navigation}

Recent research has increasingly focused on enabling legged robots to overcome the limitations of short-horizon navigation and escape local minima without prior-map information directly fed into the policy. Lee et al.~\cite{lee2024learning} develop a hierarchical RL system for wheeled-legged robots that can locally detour dead ends by feeding historically visited positions into the policy. Yang et al.~\cite{yang2025spatially} introduce a spatially enhanced recurrent network, demonstrating superior spatial memorization in maze-like environments with the current relative goal position provided to the policy at every step. REASAN~\cite{yuan2025reasan} utilizes a modular LiDAR-based framework, comprising locomotion, safety-shield, and exteroceptive estimation to facilitate effective detouring and dead-end escape in dynamic environments. Although these methods demonstrate the capability of RL to handle long-horizon navigation, they remain dependent on external state estimation for continuous relative-goal updates in real-world deployment. This dependency leaves a key part of navigation outside the learned policy, namely tracking the robot's evolving spatial relation to the goal.

\subsection{Proprioceptive Odometry}
Proprioceptive odometry provides essential ego-motion estimates for legged robot navigation in unmapped or visually degraded environments. Classical filtering methods (e.g., Extended Kalman Filters~\cite{welch1995introduction}) rely on idealized contact assumptions, making quadrupedal locomotion prone to cumulative drift from ground slippage and unmodeled terrain dynamics~\cite{bloesch2013state, maravgakis2023probabilistic}. To address these constraints, learning-based approaches directly map raw IMU and joint sequences to motion estimates, capturing complex contact behavior without explicit kinematic modeling~\cite{chen2018ionet, cioffi2023learned}. Legolas~\cite{wasserman2025legolas} learns to predict relative pose increments from proprioceptive histories entirely in simulation, achieving zero-shot transfer to real quadrupeds. AutoOdom~\cite{11375910} combines simulation pretraining with autoregressive fine-tuning on real-world data to improve odometry robustness. These results suggest that proprioceptive histories contain sufficient motion cues to infer robot egomotion and support long-horizon spatial awareness. Building on this insight, we investigate whether a navigation policy can learn to maintain this spatial context internally. While these odometry methods accumulate predicted motion increments to recover robot pose, GUIDE jointly learns direct prediction of task-relevant spatial anchors and navigation within a single-stage end-to-end framework, bringing relative-goal tracking into the learned policy without continuous external goal updates.

\begin{figure*}[!t]
    \centering
    \includegraphics[width=\textwidth]{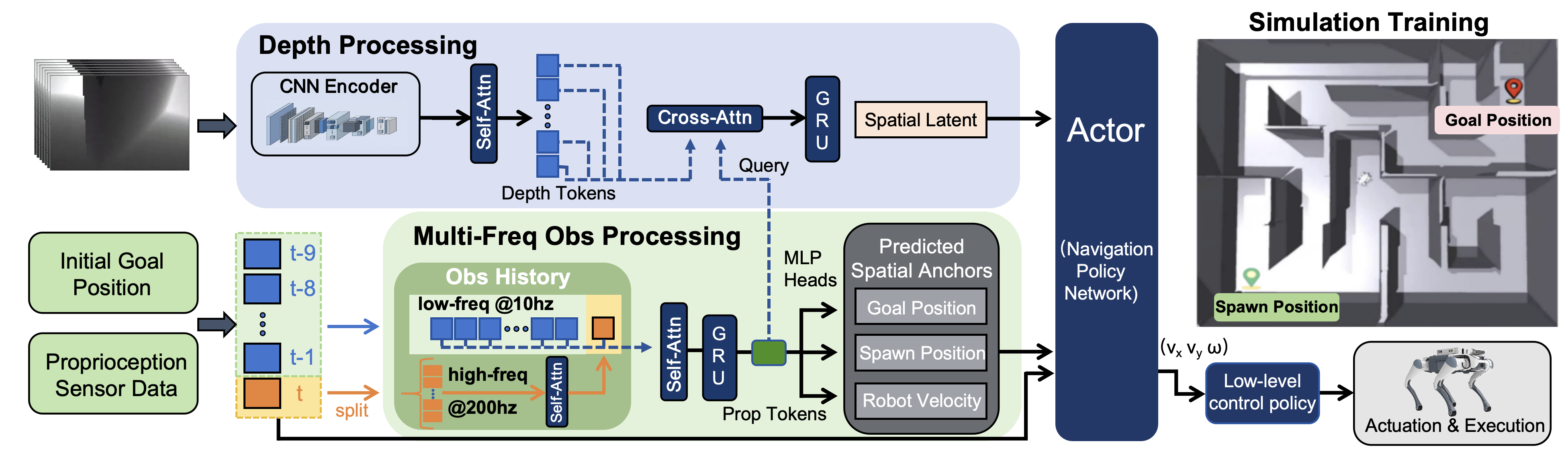}
    \vspace{-2em}
    \caption{\textbf{Overview of the GUIDE framework.}
    Multi-frequency proprioceptive history is processed into proprioceptive
    tokens, which are supervised to predict spatial anchor vectors to cultivate
    egomotion and directional awareness. Concurrently, depth buffers are encoded
    and fused with these tokens via cross-attention to yield spatial latents.
    Finally, the Actor aggregates these representations with the latest
    proprioceptive state to produce the twist commands. All navigation modules are jointly
    optimized end-to-end from scratch.}
    \label{fig:pipeline_overview}
\vspace{-2em}
\end{figure*}

\section{Method}
\label{sec:method}

\subsection{Goal-Initialized Navigation Formulation}

We formulate the navigation task as a Partially Observable Markov Decision Process (POMDP)~\cite{lauri2023partially}. At each time step $t$, the robot cannot access the global state and must make navigation decisions based on its onboard proprioception and depth inputs. In our goal-initialized setting, the robot is placed at a collision-free start position at the beginning of each episode ($t=0$) and receives an initial relative target $g^0 = [x_g^0, y_g^0, \psi_g^0]^T \in \mathbb{R}^{3}$ expressed in its egocentric frame.  Crucially, no streaming goal updates are provided in the rest of the episode ($t > 0$). To address the absence of continuous external state estimation, we introduce a spatial anchor prediction mechanism (detailed in Section \ref{subsec:guide_framework}) that is jointly trained with the navigation policy to estimate the spatial context relevant to the task $\hat{c}_t$ from the proprioceptive history. Consequently, the policy generates actions conditioned on current observation $o_t$, observation history $\mathcal{O}_t$, and the predicted spatial context $\hat{c}_t$, \emph{i.e.}, $a_t \sim \pi_{\theta}(\cdot \mid o_t, \mathcal{O}_t, \hat{c}_t)$.  The optimization objective is to learn a policy $\pi_{\theta}$ that maximizes the expected discounted return over the initial goal distribution $p(g^0)$ and the transition dynamics of the environment: $J(\pi_{\theta}) = \mathbb{E}_{\tau}\big[\sum_{t=0}^{T} \gamma^t r(s_t, a_t, g^0)\big]$, where $\tau$ denotes the induced trajectories by the policy.

\subsection{GUIDE Framework} 
\label{subsec:guide_framework}

We adopt an asymmetric actor-critic architecture optimized via Proximal Policy Optimization (PPO)~\cite{schulman2017proximal}. GUIDE follows a hierarchical control structure: the high-level navigation policy operates at $10\,\mathrm{Hz}$ to reason over multi-modal observations and produce twist commands, while a low-level locomotion controller runs at $50\,\mathrm{Hz}$ to track these commands. An overview of the framework is illustrated in Fig.~\ref{fig:pipeline_overview}.

The navigation policy takes three types of input: raw proprioceptive sensor measurements, raw depth images, and the initial target $g^0$ expressed in the robot's initial egocentric frame. The proprioceptive observations are obtained from joint encoders and the IMU. Each proprioceptive observation is a 46-dimensional vector:
\begin{equation}
p_t = [t_{\mathrm{ep}}, a_{t-1}, \omega_t, g_t^{\mathrm{grav}},
       \theta_t, \dot{\theta}_t, \theta_{t-1}^{\mathrm{cmd}}]^T,
\end{equation}
where $t_{\mathrm{ep}} \in \mathbb{R}$ denotes the elapsed time since episode initialization, and $a_{t-1} \in \mathbb{R}^{3}$ is the previously executed high-level twist command. $\omega_t \in \mathbb{R}^{3}$ is the base angular velocity, and $g_t^{\mathrm{grav}} \in \mathbb{R}^{3}$ is the projected gravity vector in the base frame. $\theta_t \in \mathbb{R}^{12}$ and $\dot{\theta}_t \in \mathbb{R}^{12}$ denote the current joint positions and joint velocities obtained from the motor encoders, respectively. Finally, $\theta^{\mathrm{cmd}}_{t-1} \in \mathbb{R}^{12}$ denotes the previous low-level joint position targets. The depth input is structured as an eight-frame history buffer $\mathcal{D}_t = \{D_{t-7}, \dots, D_t\}$, where each individual depth frame is down-sampled: $D_{t} \in \mathbb{R}^{30 \times 43}$.

\textbf{Multi-Frequency Proprioceptive Processing}
The proprioceptive inputs are encoded through two complementary temporal streams: a low-frequency history stream $\mathcal{P}_t^{\mathrm{l}} = \{p_{t-N+1}^{\mathrm{l}}, \dots, p_t^{\mathrm{l}}\}$ comprising $N=10$ samples collected at the policy rate of $10\,\mathrm{Hz}$, and a high-frequency recent stream $\mathcal{P}_t^{\mathrm{h}} = \{p_{t-M+1}^{\mathrm{h}}, \dots, p_t^{\mathrm{h}}\}$ sampled at $200\,\mathrm{Hz}$ between the previous and current policy steps, yielding $M=20$ samples per policy interval (notably $p_t^{\mathrm{h}} \equiv p_t^{\mathrm{l}}$ at current step $t$). The former tracks the historical motion context over a relatively longer temporal horizon, while the latter captures fine-grained transient dynamics. The high-frequency observations are first encoded
frame-wise, augmented with positional encodings,
and processed by self-attention. We retain the last output token
as the recent-motion embedding. This enriched token replaces the latest low-frequency token and is concatenated with the preceding $N-1$ encoded low-frequency tokens to form a unified sequence. The sequence is again augmented with positional encodings and processed by a second self-attention layer. Its last output token, integrating information from both temporal streams, is then fed into a GRU network, which maintains memory across policy steps and outputs the proprioception token $z_t^{\mathrm{p}}$.

\textbf{Spatial Anchor Prediction}
To guide the policy to extract and utilize the proprioception token, we add an auxiliary supervision signal through a spatial anchor predictor to the token $z_t^{\mathrm{p}}$. Unlike the typical odometry-based pipeline that integrates incremental motion estimates~\cite{wasserman2025legolas,11375910} and derives task states from the resulting pose, this predictor employs three lightweight MLP heads to directly estimate the robot's body-frame velocity $\hat{v}_t$,
relative goal position $\hat{g}_t$, and relative spawn position
$\hat{s}_t$. During training, these predictions are supervised using privileged simulation states, where the estimation loss is defined as: \vspace{-0.2em}
\begin{equation}
    \mathcal{L}_{\mathrm{est}} = \|\hat{v}_t - v_t\|_2^2 + \mathcal{H}_\delta(\hat{g}_t - g_t) + \mathcal{H}_\delta(\hat{s}_t - s_t),
\end{equation}  
\vspace{-0.2em}
Here, $v_t$, $g_t$, and $s_t$ denote the corresponding ground-truth
simulation states, and $\mathcal{H}_{\delta}(\cdot)$ denotes the smooth
L1 (Huber) loss, with $\delta=1$ for both position terms.

\textbf{Depth Processing}
 Each individual frame within $\mathcal{D}_t$ is independently processed by a shared convolutional encoder to extract depth features. These frame-wise features are then integrated through a self-attention layer to fuse the temporal information across the buffer, producing the depth tokens $z_t^{\mathrm{d}}$. To condition the depth features on the robot's motion state, a cross-attention module is employed using the proprioceptive token $z_t^{\mathrm{p}}$ as the query and the depth tokens $z_t^{\mathrm{d}}$ as keys and values. The resulting cross-modal features are passed through a GRU layer to produce the final spatial latent feature $z_t^{\mathrm{s}}$.  

Ultimately, the navigation policy network takes the concatenated vector $[z_t^{\mathrm{s}}, \hat{v}_t, \hat{g}_t, \hat{s}_t, p_t^{\mathrm{l}}]$ as input and outputs the egocentric twist commands $a_t = [v_x, v_y, \omega_z]^T$, where $p_t^{\mathrm{l}}$ denotes the latest frame of proprioceptive observations. The commands are subsequently executed on the robot by the low-level locomotion controller, which is trained following the framework established by Long et al.~\cite{long2024hybrid}.

\subsection{Reward Design and Multi-Critic Value Estimation}
\begin{table}[!t]
\centering
\footnotesize
\setlength{\tabcolsep}{2pt}
\caption{Complete reward terms and coefficients.}
\label{tab:reward_details}
\begin{tabularx}{\linewidth}{@{}>{\raggedright\arraybackslash}p{0.27\linewidth} >{\centering\arraybackslash}X r@{}}
\toprule
\textbf{Reward Term} & \textbf{Formulation} & \textbf{Coeff.} \\
\midrule
\multicolumn{3}{@{}l}{\textbf{Sparse task rewards}} \\
Position reaching
& $\mathbb{I}_{\mathrm{complete}} /
  (1 + (e_t^{\mathrm{pos}}/\sigma_{\mathrm{pos}})^2)$
& $3.0$ \\

Heading reaching
& $\mathbb{I}_{\mathrm{complete}} /
  (1 + (e_t^{\mathrm{head}}/\sigma_{\mathrm{head}})^2)$
& $1.5$ \\
Bad termination
& $\mathbb{I}_{\mathrm{bad}}$
& $-1.0$ \\

\midrule
\multicolumn{3}{@{}l}{\textbf{Dense shaping rewards}} \\
Euclidean progress
& $\frac{\max(0, d_{\mathrm{euc}}^{\mathrm{best}} - d_t^{\mathrm{euc}})}{d_{t=0}^{\mathrm{euc}}}$
& $60.0$ \\
Maze-cell progress
& $\frac{\max(0, d_{\mathrm{cell}}^{\mathrm{best}} - d_t^{\mathrm{cell}})}{d_{t=0}^{\mathrm{cell}}}$
& $60.0$ \\
Near-goal velocity damping
& $\exp(-\sum_i |a_{t,i}| / a_i^{\max})\mathbb{I}_{\mathrm{soft}}$
& $5.0$ \\
No-motion penalty
& $\mathbb{I}_{\mathrm{static}}$
& $-5.0$ \\
Collision penalty
& $\mathbb{I}_{\mathrm{coll}}$
& $-50.0$ \\
Command overspeed penalty
& $\sum_i \max(|a^{\mathrm{raw}}_{t,i}| - \lambda a_i^{\max}, 0)^2$
& $-2.5$ \\
Command rate
& $\|a_t - a_{t-1}\|_2^2$
& $-0.05$ \\
Joint acceleration
& $\|\ddot{\theta}_t\|_2^2$
& $-2.5 \times 10^{-7}$ \\
Joint torque
& $\|\tau_t\|_2^2$
& $-2.0 \times 10^{-4}$ \\
Linear velocity (z)
& $v_{t,z}^2$
& $-0.1$ \\
Roll-pitch angular velocity
& $\|\omega_{t,xy}\|_2^2$
& $-0.005$ \\
Orientation
& $\|g_{t,xy}^{\mathrm{grav}}\|_2^2$
& $-0.1$ \\
\bottomrule
\end{tabularx}
\par\smallskip
\raggedright
Parameters: $\sigma_{\mathrm{pos}}=0.2\,\mathrm{m}$,
$\sigma_{\mathrm{head}}=0.5\,\mathrm{rad}$,
$\sigma_{\mathrm{soft}}=2.0\,\mathrm{m}$, $\lambda=1.3$.
\vspace{-2em}
\end{table}

\textbf{Reward Function Design}
Our reward design combines sparse task-level rewards with dense progress rewards. The complete list of reward terms is provided in Table~\ref{tab:reward_details}. Task-level rewards ($r_{\mathrm{pos}}$, $r_{\mathrm{head}}$) follow the formulation of $r_{\mathrm{track}}$, while dense progress rewards ($r_{\mathrm{prog}}^{\mathrm{euc}}$, $r_{\mathrm{prog}}^{\mathrm{cell}}$) follow the formulation of $r_{\mathrm{prog}}$:
\vspace{-0.2em}
\begin{equation}
    r_{\mathrm{track}} = \frac{\mathbb{I}_{\mathrm{complete}}}{1 + (e / \sigma)^2},\qquad
    r_{\mathrm{prog}} = \frac{\max(0, d^{\mathrm{best}} - d_t)}{d_{t=0}},
\end{equation}  

where $e$ and $\sigma$ denote the corresponding tracking error and
scale, and $\mathbb{I}_{\mathrm{complete}}$ activates only when the relative goal distance $d_t$ falls below $\sigma_{\mathrm{pos}}$.
$\mathbb{I}_{\mathrm{bad}}$ indicates failure termination when the
robot falls over. $\mathbb{I}_{\mathrm{soft}}$ activates when the goal is within $\sigma_{\mathrm{soft}}$ and no obstacle lies between the robot and the goal. For progress terms, $d_t$ and $d^{\mathrm{best}}$ denote the current and previous minimum distances, and progress is normalized by the initial distance $d_{t=0}$. Cluttered environments use the Euclidean progress reward $r_{\mathrm{prog}}^{\mathrm{euc}}$, with $d$ denoting the straight-line distance to the goal. Maze environments use $r_{\mathrm{prog}}^{\mathrm{cell}}$, with $d$ denoting the maze-graph connectivity distance to the goal.

\begin{figure}[!t]
\centering
\includegraphics[width=\linewidth]{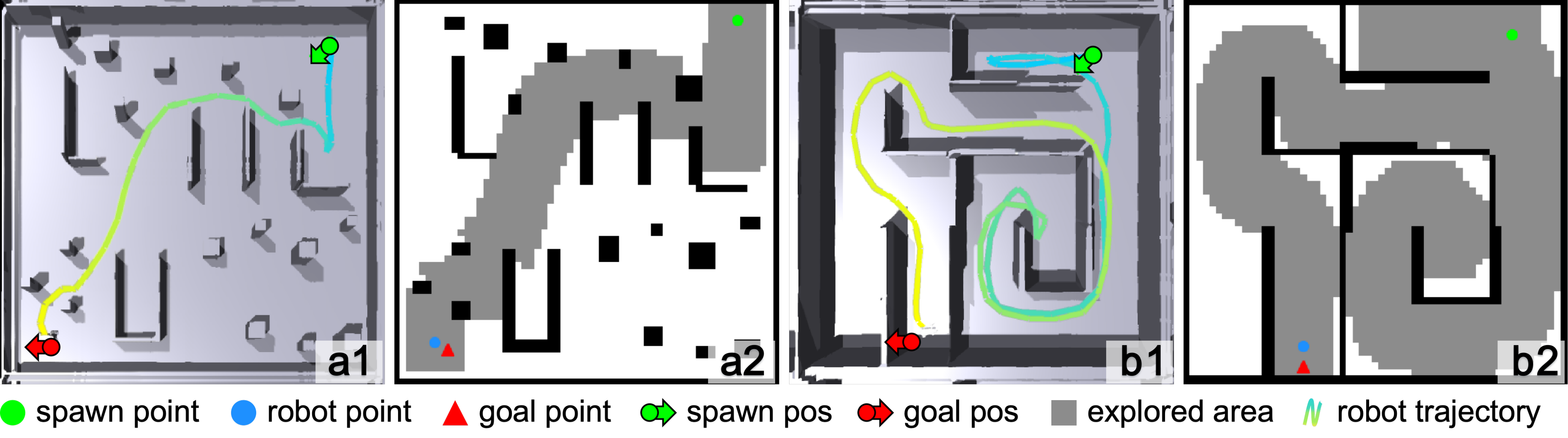}
\caption{\textbf{(a1, b1):} Cluttered and maze terrains, with curves in gradient color indicating the robot's trajectories. \textbf{(a2, b2):} Overlaid visualizations of the occupancy and explored-region maps, where black denotes obstacles, gray denotes explored free space, and white denotes unexplored free space. The two maps are stacked as separate input channels for the privileged critic.}
\label{fig:terrain}
\vspace{-2em}
\end{figure}

\textbf{Privileged Critic Learning}
Under the asymmetric architecture, the value networks leverage privileged information from simulation to improve value estimation and guide policy optimization. As shown in Fig.~\ref{fig:terrain}, we provide the critic with a downsampled top-down occupancy map $\mathcal{M}_{\mathrm{occ}}$ and an explored-region map $\mathcal{M}_{\mathrm{exp}}$, thereby providing the critic with explicit global information about the environmental geometry and the robot's traversal trajectory. These maps are stacked as a two-channel tensor and processed by a dedicated convolutional network. 
\textbf{Multi-Critic Paradigm:} To better estimate the mixture of dense and sparse rewards with distinct temporal characteristics, we introduce a Multi-Critic (MuC) architecture~\cite{mysore2022multi, wang2026puma}. The total reward is partitioned into a sparse task group $r_t^{\mathrm{s}}$ and a dense shaping group $r_t^{\mathrm{d}}$. The MuC module outputs independent value estimates $V_{\phi}^{k}(s_t)$ for $k \in \{\mathrm{s}, \mathrm{d}\}$, optimized against their bootstrapped returns $\hat{R}_t^{k}$ by minimizing $\mathcal{L}_{\mathrm{value}} = \sum_{k \in \{\mathrm{s}, \mathrm{d}\}} \mathbb{E}[(\hat{R}_t^{k} - V_{\phi}^{k}(s_t))^2]$. We then compute the unified normalized advantage $\hat{A}_t$ from the individual generalized advantage estimates (GAE)~\cite{Schulman2015HighDimensionalCC} $\hat{A}_t^k$:

\begin{equation}
    \hat{A}_t = \big( \sum_{k \in \{\mathrm{s}, \mathrm{d}\}} w_k \hat{A}_t^k - \mu_{\mathrm{MuC}} \big) \big/ \sigma_{\mathrm{MuC}},
\end{equation}

where $w_k$ denotes the reward-group weight, while $\mu_{\mathrm{MuC}}$ and $\sigma_{\mathrm{MuC}}$ are the batch mean and standard deviation of the combined advantages. 

\subsection{Training Details}

\textbf{Terrain Design.}
To train a policy with both reactive obstacle avoidance and long-horizon reasoning capabilities, we procedurally generate two distinct terrain types. As illustrated in Fig.~\ref{fig:terrain}, \textbf{Cluttered terrains} are constructed by randomly placing specific obstacle categories, including narrow alleys, L-shaped corners, and pillars. The resulting dense and irregular geometry creates environments with frequent occlusions and local concave traps. \textbf{Maze terrains} are grid-based perfect mazes generated by randomized depth-first search (DFS)~\cite{tarjan1972depth}. These loop-free structures guarantee a unique path between any two grid cells while providing scalable topological difficulty through branching corridors and dead ends, requiring the robot to retrace its route after entering an incorrect branch.  

\textbf{Training Curriculum.}
We adopt a curriculum that progressively increases both terrain complexity and initial robot-to-goal distance. Specifically, terrain complexity is increased by raising obstacle density in cluttered environments and increasing the number of grid cells in mazes. The initial robot-to-goal distance is progressively increased, measured by Euclidean distance in cluttered environments and maze-graph connectivity distance in mazes. More precisely, the latter counts the transitions between adjacent cells along the unique traversable path from the start cell to the goal cell. 

\textbf{Simulation Setup.} We train the 12-DoF Deep-Robotics
Lite3 robot using Isaac Gym across 4,096 parallel environments on a single NVIDIA H100 GPU. During training, we randomize dynamics parameters and inject noise into proprioceptive sensor
measurements and depth images. The entire navigation reinforcement learning process is completed end-to-end in a single unified phase without any pretraining, with the policy typically converging within 20,000 iterations.

\section{Experimental Results}
\label{sec:result}

\subsection{Simulation Benchmarks and Ablations}
To assess the contributions of GUIDE's components and compare against baseline methods, we consider the following configurations. First, for internal component ablations:
(i) \textbf{w/o High-Frequency Prop} removes the $200\,\mathrm{Hz}$ high-frequency proprioceptive observation branch, restricting the policy to utilize only the $10\,\mathrm{Hz}$ observation history processed via self-attention; (ii) \textbf{w/o Target Prediction}, (iii) \textbf{w/o Spawn Prediction}, and (iv) \textbf{w/o Velocity Prediction} deactivate their respective prediction heads; (v) \textbf{w/o Multi-Critic} replaces the multi-critic architecture with a standard single-critic network to predict the total return. 
Second, external baselines: 
(vi) \textbf{GUIDE-Legolas} replaces GUIDE's spatial-anchor predictor with the pretrained Legolas odometry model~\cite{wasserman2025legolas}, deriving body-frame velocity from its predicted relative pose increments and accumulating these increments to compute the current relative goal and spawn positions;
(vii) \textbf{SRU}~\cite{yang2025spatially}, which is trained with continuous streaming ground-truth relative-goal updates.
Finally, to investigate performance boundaries, we evaluate two ground-truth oracle variants: (viii) \textbf{GUIDE-GT-Inference}, which directly replaces the predicted goal and spawn positions with ground-truth values from the simulation in test experiments. (ix) \textbf{GUIDE-GT-Train}, which completely removes the spatial-anchor predictor and is trained from scratch, receiving ground-truth relative goal and spawn positions at every control step during both training and inference.

\begin{figure}[tbp] %
  \centering
  \includegraphics[width=\linewidth]{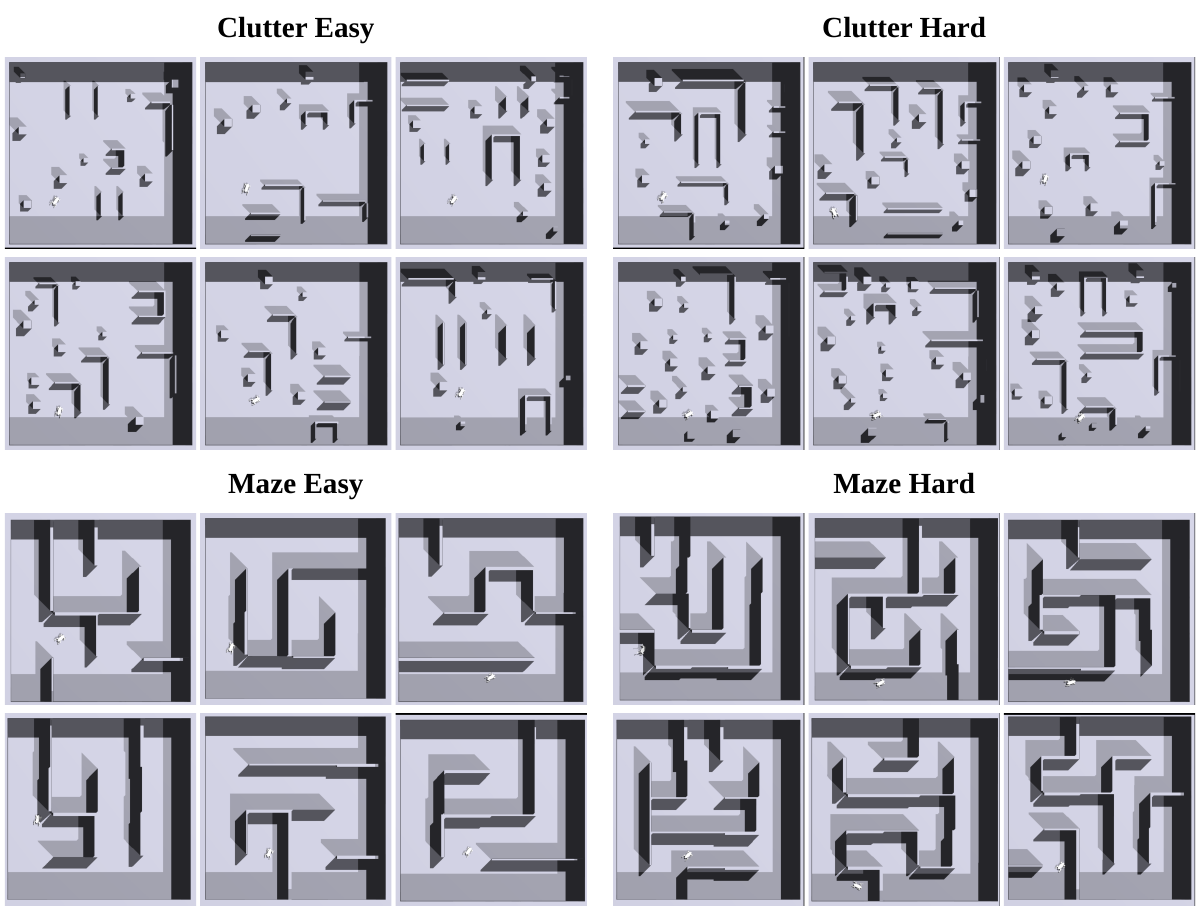} 
   \vspace{-1.5em}
  \caption{\textbf{Example simulation benchmark layouts.}
Benchmark difficulty is determined by obstacle density in cluttered terrains
and the number of grid cells in maze terrains. Each evaluation run uses 50 distinct layouts per setting, generated using randomized seeds.}
  \label{fig:benchmark_terrains}
\vspace{-2em}
\end{figure}
\label{subsec:sim_experiments}
We evaluate the compared policies across cluttered and maze terrains at two difficulty levels, resulting in four benchmark settings shown in Fig.~\ref{fig:benchmark_terrains}. Except for SRU and the oracle variants, all methods receive the goal only at initialization. Performance is quantified by three key metrics: (i) \textbf{Success Rate (SR)}, computed as the percentage of trials reaching within a $0.2$m radius of the goal; (ii) \textbf{Arrival Time (AT)}, defined as the mean episode duration across all trials, subject to a $60\,\mathrm{s}$ timeout; and (iii) \textbf{Collision Rate (CR)}, defined as the percentage of episodes with at least one physical collision. For each benchmark, we perform 10 evaluation runs, each comprising $1,000$ episodes across $50$ layouts generated using a different set of random seeds. Table~\ref{tab:main_results} reports the mean and standard deviation across these runs.

\begin{table*}[tbp]
\centering
\footnotesize
\setlength{\tabcolsep}{0pt}

\def\unc#1#2{%
    \mbox{#1{\scriptsize\color{black!55}$\pm$#2}}%
}

\caption{Quantitative comparison across four simulation benchmark settings.}
\label{tab:main_results}
\vspace{-0.5em}
\begin{tabular*}{\textwidth}{@{\extracolsep{\fill}} l ccc ccc ccc ccc @{}}
\toprule
& \multicolumn{6}{c}{\textbf{Cluttered Environments}}
& \multicolumn{6}{c}{\textbf{Maze Environments}} \\
\cmidrule(lr){2-7} \cmidrule(lr){8-13}
& \multicolumn{3}{c}{Easy}
& \multicolumn{3}{c}{Hard}
& \multicolumn{3}{c}{Easy}
& \multicolumn{3}{c}{Hard} \\
\cmidrule(lr){2-4} \cmidrule(lr){5-7}
\cmidrule(lr){8-10} \cmidrule(lr){11-13}
\textbf{Policy Variant}
& SR$\uparrow$ & AT$\downarrow$ & CR$\downarrow$
& SR$\uparrow$ & AT$\downarrow$ & CR$\downarrow$
& SR$\uparrow$ & AT$\downarrow$ & CR$\downarrow$
& SR$\uparrow$ & AT$\downarrow$ & CR$\downarrow$ \\
& [\%] & [s] & [\%]
& [\%] & [s] & [\%]
& [\%] & [s] & [\%]
& [\%] & [s] & [\%] \\
\midrule

\textbf{GUIDE (Full)}
& \unc{99.9}{0.1} & \unc{6.85}{0.8} & \unc{3.20}{0.6}
& \unc{99.9}{0.1} & \unc{7.44}{1.0} & \unc{4.40}{0.8}
& \unc{99.9}{0.1} & \unc{15.7}{0.9} & \unc{3.40}{0.3}
& \unc{99.6}{0.3} & \unc{22.1}{0.9} & \unc{4.80}{0.4} \\
\midrule

(i) w/o High-Freq. Prop
& \unc{0.16}{0.2} & \unc{57.9}{1.8} & \unc{12.8}{1.7}
& \unc{0.08}{0.1} & \unc{58.6}{1.2} & \unc{13.3}{3.1}
& \unc{0.06}{0.1} & \unc{59.2}{0.6} & \unc{2.80}{0.4}
& \unc{0.01}{0.0} & \unc{59.8}{0.1} & \unc{3.10}{0.4} \\

(ii) w/o Target Prediction
& \unc{97.8}{0.4} & \unc{8.22}{0.9} & \unc{6.40}{1.2}
& \unc{96.7}{0.6} & \unc{8.86}{0.9} & \unc{8.20}{1.5}
& \unc{98.2}{0.4} & \unc{18.2}{1.3} & \unc{6.20}{0.6}
& \unc{70.9}{3.6} & \unc{34.9}{2.4} & \unc{8.40}{0.9} \\

(iii) w/o Spawn Prediction
& \unc{98.8}{0.6} & \unc{7.50}{1.1} & \unc{4.40}{0.9}
& \unc{98.2}{0.5} & \unc{7.66}{1.1} & \unc{4.80}{1.4}
& \unc{98.9}{0.4} & \unc{16.5}{1.5} & \unc{4.90}{0.7}
& \unc{90.1}{2.1} & \unc{26.7}{2.8} & \unc{6.60}{1.3} \\

(iv) w/o Velocity Prediction
& \unc{99.0}{0.3} & \unc{8.84}{1.7} & \unc{7.40}{2.1}
& \unc{98.5}{0.5} & \unc{9.50}{2.0} & \unc{12.2}{4.3}
& \unc{96.3}{2.5} & \unc{19.9}{2.4} & \unc{7.80}{1.9}
& \unc{53.5}{7.2} & \unc{39.4}{4.5} & \unc{11.30}{3.6} \\

(v) w/o Multi-Critic
& \unc{95.2}{2.1} & \unc{10.6}{1.4} & \unc{13.3}{4.3}
& \unc{88.5}{2.7} & \unc{11.8}{2.1} & \unc{15.4}{5.5}
& \unc{82.0}{3.4} & \unc{26.9}{2.9} & \unc{6.50}{2.3}
& \unc{49.1}{6.8} & \unc{41.0}{4.3} & \unc{5.40}{2.1} \\
\midrule
(vi) GUIDE-Legolas ~\cite{wasserman2025legolas}
& \unc{82.5}{3.1} & \unc{17.6}{2.1} & \unc{5.0}{1.2}
& \unc{77.6}{4.3} & \unc{20.6}{3.2} & \unc{6.3}{1.6}
& \unc{76.9}{3.5} & \unc{28.5}{2.3} & \unc{5.50}{1.3}
& \unc{63.3}{5.4} & \unc{39.8}{3.0} & \unc{6.40}{1.6} \\

(vii) SRU~\cite{yang2025spatially}
& \unc{96.2}{0.7} & \unc{10.3}{0.5} & \unc{8.04}{2.1}
& \unc{93.6}{1.0} & \unc{12.8}{0.7} & \unc{9.65}{4.0}
& \unc{85.3}{2.8} & \unc{24.2}{1.2} & \unc{5.24}{0.9}
& \unc{76.3}{4.0} & \unc{41.6}{1.7} & \unc{5.80}{1.2} \\
\midrule
(viii) GUIDE-GT-Inference
& \unc{100.0}{0.0} & \unc{6.47}{0.7} & \unc{2.60}{0.6}
& \unc{99.9}{0.1} & \unc{6.54}{0.9} & \unc{3.10}{0.8}
& \unc{99.9}{0.1} & \unc{15.2}{0.9} & \unc{2.40}{0.2}
& \unc{99.9}{0.1} & \unc{20.7}{1.0} & \unc{3.60}{0.5} \\

(ix) GUIDE-GT-Train
& \unc{100.0}{0.0} & \unc{6.43}{0.7} & \unc{3.55}{0.6}
& \unc{100.0}{0.0} & \unc{6.69}{0.8} & \unc{5.62}{0.9}
& \unc{99.9}{0.1} & \unc{14.8}{0.3} & \unc{4.24}{0.5}
& \unc{99.8}{0.1} & \unc{20.6}{0.7} & \unc{6.02}{0.7} \\

\bottomrule
\end{tabular*}

\end{table*}

Beyond task performance, we further evaluate the spatial anchor prediction accuracy to understand how the supervised predictor contributes to the ablation results. Specifically, this analysis examines the components that shape the spatial anchor representation, including high-frequency proprioceptive input and individual prediction heads, while also comparing GUIDE's spatial estimation accuracy with Legolas~\cite{wasserman2025legolas}. Under the same benchmark settings, we measure the mean tracking error (MTE) of the relative target position, the relative spawn position, and the body velocity, averaged across all policy steps within the evaluation episodes. The corresponding results and standard deviation  are shown in Table~\ref{tab:state_estimation}. 
Based on these quantitative results, we derive the following key insights.

\begin{table*}[tbp]
\centering
\footnotesize
\setlength{\tabcolsep}{0pt}

\def\errstd#1#2{%
    \mbox{#1%
    \if\relax\detokenize{#2}\relax
    \else
        {\scriptsize\color{black!55}$\pm$#2}%
    \fi}%
}

\caption{Spatial anchor prediction error across four benchmark settings
in simulation.}
\label{tab:state_estimation}

\begin{tabular*}{\textwidth}{@{\extracolsep{\fill}} l ccc ccc ccc ccc @{}}
\toprule
& \multicolumn{6}{c}{\textbf{Cluttered Environments}}
& \multicolumn{6}{c}{\textbf{Maze Environments}} \\
\cmidrule(lr){2-7} \cmidrule(lr){8-13}
& \multicolumn{3}{c}{Easy}
& \multicolumn{3}{c}{Hard}
& \multicolumn{3}{c}{Easy}
& \multicolumn{3}{c}{Hard} \\
\cmidrule(lr){2-4} \cmidrule(lr){5-7}
\cmidrule(lr){8-10} \cmidrule(lr){11-13}
\textbf{Policy Variant}
& Tgt. & Spn. & Vel.
& Tgt. & Spn. & Vel.
& Tgt. & Spn. & Vel.
& Tgt. & Spn. & Vel. \\
& [m] & [m] & [m/s]
& [m] & [m] & [m/s]
& [m] & [m] & [m/s]
& [m] & [m] & [m/s] \\
\midrule

\textbf{GUIDE (Full)}
& \errstd{0.21}{0.09} & \errstd{0.17}{0.09} & \errstd{0.12}{0.07}
& \errstd{0.23}{0.10} & \errstd{0.18}{0.09} & \errstd{0.12}{0.07}
& \errstd{0.22}{0.12} & \errstd{0.18}{0.12} & \errstd{0.11}{0.07}
& \errstd{0.31}{0.17} & \errstd{0.27}{0.15} & \errstd{0.11}{0.08} \\
\midrule

w/o High-Freq. Prop
& \errstd{3.43}{2.72} & \errstd{2.92}{2.15} & \errstd{0.18}{0.12}
& \errstd{3.37}{2.85} & \errstd{2.84}{2.26} & \errstd{0.18}{0.13}
& \errstd{3.10}{2.34} & \errstd{2.82}{2.14} & \errstd{0.10}{0.09}
& \errstd{3.16}{2.55} & \errstd{2.76}{2.34} & \errstd{0.11}{0.09} \\

w/o Target Prediction
& -- & \errstd{0.13}{0.07} & \errstd{0.11}{0.07}
& -- & \errstd{0.17}{0.08} & \errstd{0.13}{0.09}
& -- & \errstd{0.32}{0.17} & \errstd{0.11}{0.07}
& -- & \errstd{0.54}{0.35} & \errstd{0.10}{0.07} \\

w/o Spawn Prediction
& \errstd{0.34}{0.15} & -- & \errstd{0.12}{0.08}
& \errstd{0.33}{0.18} & -- & \errstd{0.12}{0.08}
& \errstd{0.38}{0.20} & -- & \errstd{0.11}{0.07}
& \errstd{0.46}{0.25} & -- & \errstd{0.11}{0.08} \\

w/o Velocity Prediction
& \errstd{0.29}{0.14} & \errstd{0.18}{0.12} & --
& \errstd{0.29}{0.17} & \errstd{0.19}{0.15} & --
& \errstd{0.37}{0.21} & \errstd{0.33}{0.20} & --
& \errstd{0.67}{0.37} & \errstd{0.65}{0.44} & -- \\

\midrule

GUIDE-Legolas~\cite{wasserman2025legolas}
& \errstd{0.48}{0.36} & \errstd{0.51}{0.66} & \errstd{0.07}{0.04}
& \errstd{0.51}{0.43} & \errstd{0.53}{0.67} & \errstd{0.07}{0.04}
& \errstd{0.40}{0.28} & \errstd{0.33}{0.28} & \errstd{0.07}{0.03}
& \errstd{0.44}{0.30} & \errstd{0.42}{0.35} & \errstd{0.07}{0.03} \\

\bottomrule
\end{tabular*}
\vspace{-2em}
\end{table*}

\textbf{High-Frequency Proprioception is Crucial for Egomotion and Directional Awareness.} The \textbf{w/o High-Frequency Prop} variant suffers a severe performance collapse across all tasks, almost completely failing to reach the target within the time limit. Instead of actively navigating, the robot quickly becomes trapped in local dead ends, which explains its low collision rate in the maze environment. As revealed in Table~\ref{tab:state_estimation}, the policy entirely loses its internal tracking of the target and spawn positions, alongside degraded velocity estimation in cluttered environments, which suggests that maintaining a reliable sense of egomotion and direction under partial observability strongly relies on the fine-grained motion information provided by high-frequency proprioceptive observations. This capability is particularly important in our goal-initialized setting, where the navigation policy must learn to maintain spatial awareness internally without continuous external relative-goal updates.

\textbf{Spatial Anchor Predictions Jointly Shape Spatial Context.}
GUIDE consistently outperforms all variants lacking individual prediction heads. Notably, \textbf{w/o Target Prediction} suffers a larger decline in task performance than \textbf{w/o Spawn Prediction}, as shown in Table~\ref{tab:main_results}. We attribute this gap to the more direct goal-oriented guidance provided by the target position, while the spawn prediction requires the policy to implicitly infer the spatial relation between the start and the target. Among the three prediction-head ablations, \textbf{w/o Velocity Prediction} experiences the most severe degradation in both success rate and collision avoidance. These results indicate that robust navigation requires both instantaneous motion cues from velocity prediction and long-horizon positional anchors from target and spawn prediction, which together help the policy construct a coherent spatial context.

\textbf{GUIDE Effectively Exploits Learned Spatial Context for Robust Navigation.}
GUIDE achieves higher SR than SRU~\cite{yang2025spatially} across all four benchmarks. In observed SRU failures, the policy repeatedly cycles between neighboring dead ends, whereas GUIDE backtracks and escapes. GUIDE-Legolas yields lower instantaneous velocity error. However, accumulating its frame-to-frame pose estimates to recover the robot pose and transform the goal and spawn positions into the current robot frame results in larger position errors and lower SR. This highlights the advantage of our proposed method that directly learns task-relevant spatial anchors jointly with the navigation policy. Without prediction errors in navigation process, both oracle variants (\textbf{GUIDE-GT-Inference} and \textbf{GUIDE-GT-Train}) achieve slightly better performance than \textbf{GUIDE} in terms of success rate and arrival time. Although this improvement reveals the remaining limitation of the learned internal navigation states, the small gap also indicates that GUIDE effectively learns and exploits such spatial context from partial observations to accomplish long-horizon navigation. This supports the central hypothesis of GUIDE: an end-to-end RL policy can maintain internal egomotion and directional awareness and successfully navigate complex environments when the target is provided only at initialization. 
Additionally, the significant performance drop in \textbf{w/o Multi-Critic}  demonstrates that decoupling the value estimation for dense and sparse rewards is critical for policy learning.

\subsection{Real-World Experiments}
\label{subsec:real_world}

We deploy our framework on a Deep-Robotics Lite3 quadruped robot. The policy runs onboard on an NVIDIA Jetson Orin NX, using proprioceptive measurements from the robot and depth images from an Intel RealSense D435i camera. The depth stream is processed
with the official RealSense filtering pipeline, updating depth frames at 10 Hz. For real-world quantitative evaluation, we use a Livox MID360 LiDAR with Faster-LIO~\cite{bai2022fasterlio} for mapping and robot pose estimation. This LiDAR-based pipeline is used solely for evaluation and provides no input to the navigation policy.
\begin{figure*}[!t]
\centering
\includegraphics[width=\textwidth]{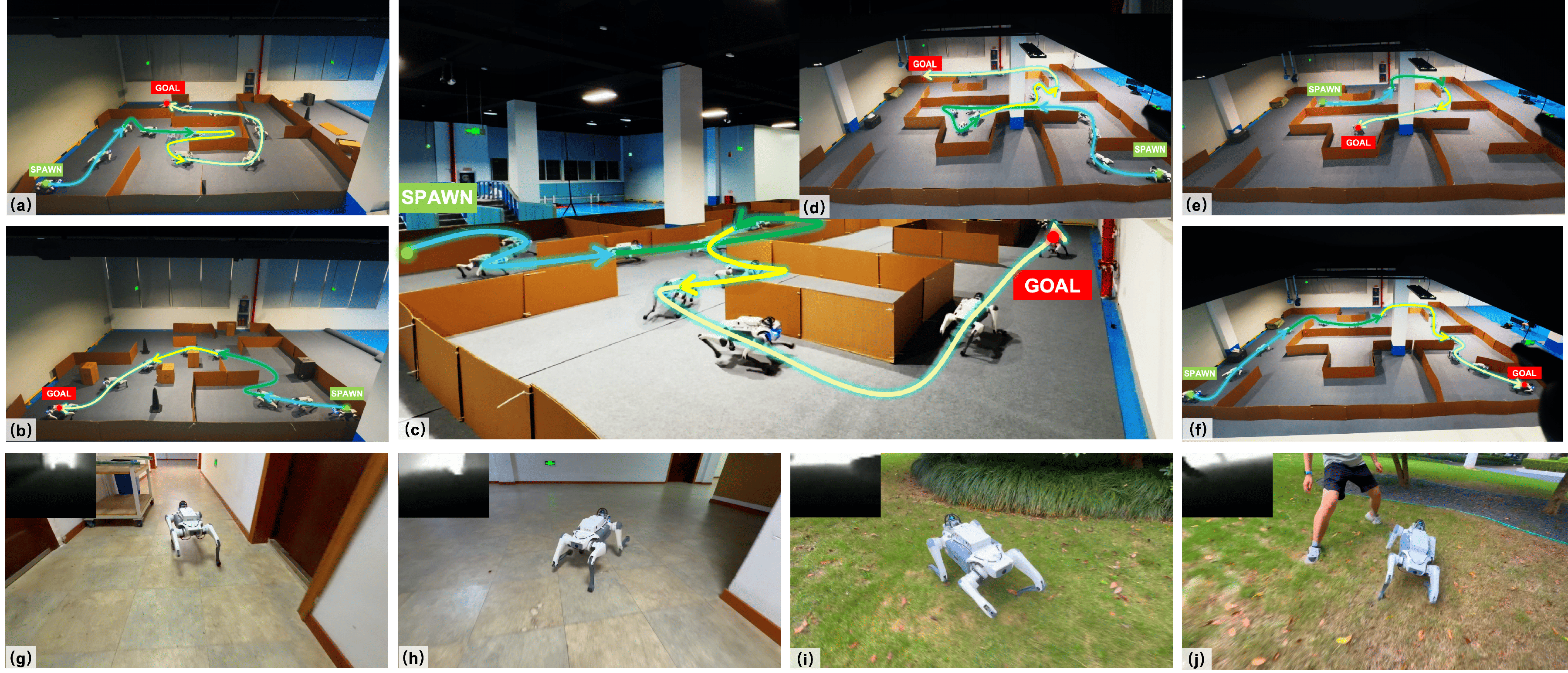}
\caption{\textbf{Real-world deployments.}
GUIDE navigates through \textbf{(a)} maze and
\textbf{(b)} cluttered environments.
\textbf{(c)} illustrates escape from multiple dead ends during a representative navigation trial, with \textbf{(d)} providing an overhead view of the same trial. \textbf{(e, f)} show navigation under different start--goal configurations. Additional deployments include \textbf{(g, h)} narrow office corridors and outdoor grasslands with \textbf{(i)} dense vegetation and \textbf{(j)} dynamic obstacles.
Colored curves in (a--f) indicate executed trajectories,
while top-left insets in (g--j) show onboard depth observations.}

\label{fig:real_demo}
\vspace{-1.5em}
\end{figure*}
\begin{figure}[!htbp] %
  \centering   
  \includegraphics[width=\linewidth]{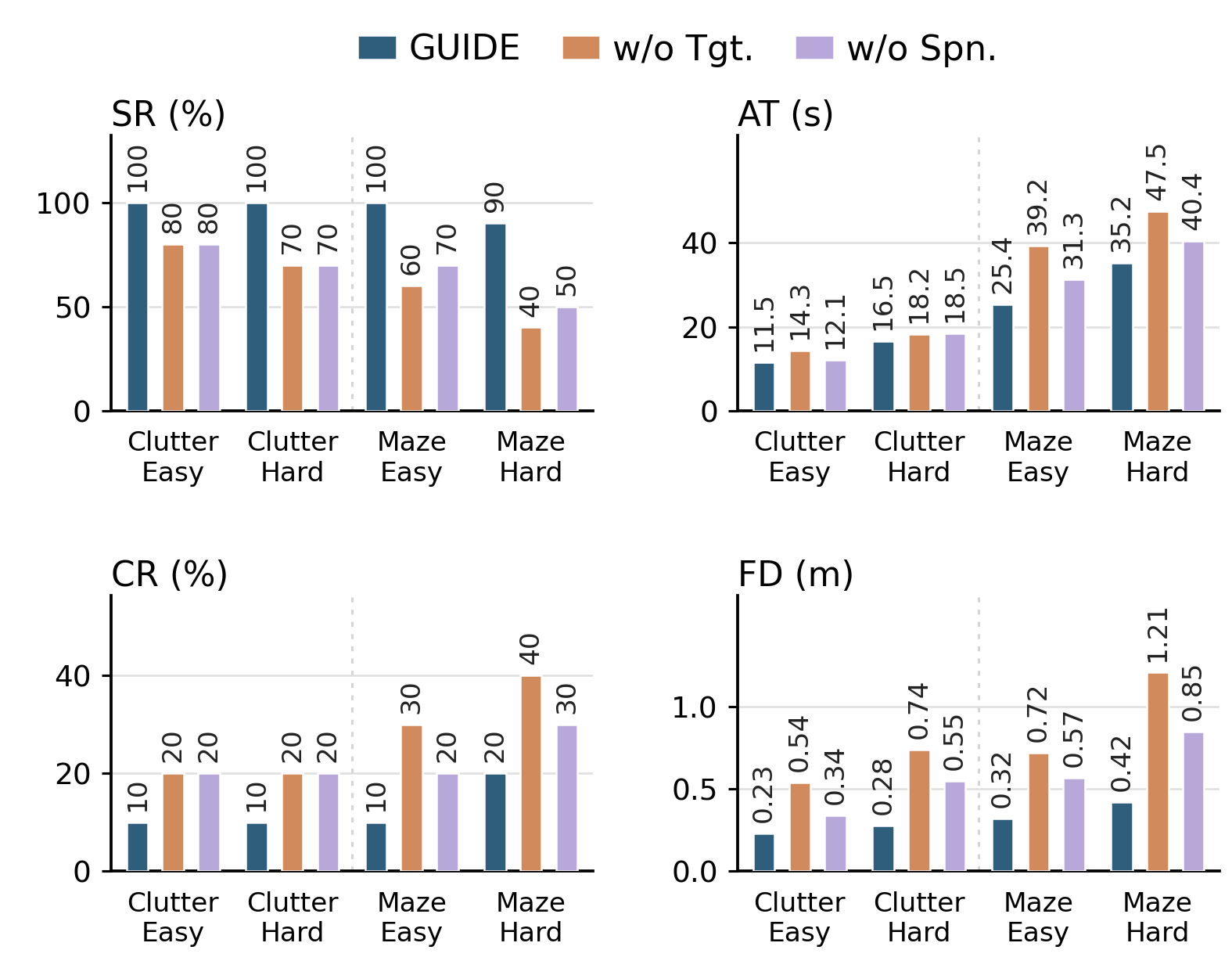} 
  \vspace{-1.5em}
  \caption{\textbf{Quantitative results of real-world environments.} \textbf{GUIDE} achieves the best results across all metrics. The robot autonomously stops when its internally estimated distance to the goal falls below $0.2$ m. Therefore, the resulting average final distances \textbf{(FDs)} of approximately $0.2–0.3$ m indicate that the learned internal goal representation remains reasonably accurate during real-world deployment, supporting the zero-shot sim-to-real transfer of the policy.}
  \label{fig:real_world_results}
   \vspace{-2em}
\end{figure}

\vspace{-1em}
\textbf{In-Lab Evaluations:} We evaluate GUIDE in cluttered and maze environments, each with Easy and Hard settings corresponding to those used in simulation. Representative deployments are shown in Fig.~\ref{fig:real_demo}. The longest executed trajectories reach $24.5\,\mathrm{m}$
and $56.1\,\mathrm{m}$ in the Easy and Hard settings, respectively. During physical deployment, the robot is programmed to stop autonomously when its internally estimated distance to the target falls below $0.2\,\mathrm{m}$. To quantify the accuracy of the learned internal directional awareness, we introduce an additional metric: \textbf{Final Distance (FD)}, defined as the actual physical distance between the robot's base and the goal location when the robot ``thinks" it has reached the goal based on its internal direction awareness. For physical deployment, we evaluate GUIDE and the two ablation variants with the highest simulation success rates. Each method is evaluated over 10 independent physical trials per setting, with the quantitative results detailed in Fig.~\ref{fig:real_world_results}.  Across all scenarios, GUIDE achieves the best overall performance and reaches the target with the highest precision, demonstrating the robustness of our framework.

\textbf{In-the-wild Deployments:} We further deploy GUIDE in unstructured environments unseen during training. As illustrated in Fig.~\ref{fig:real_demo}, these scenarios range from long office corridors to outdoor grasslands, where the robot must dynamically avoid obstacles and detour around dense vegetation. Despite drastic changes in environment appearance, physical terrain attributes, lighting, and obstacle shapes, GUIDE maintains stable navigation behavior using solely raw onboard sensing, achieving successful zero-shot sim-to-real transfer and overcoming real-world domain shifts.

\section{Conclusion}
\label{sec:conclusion}

In this work, we presented GUIDE, an end-to-end reinforcement learning framework for goal-initialized legged navigation. By utilizing multi-frequency raw proprioception and raw depth information, GUIDE enables legged robots to cultivate internal egomotion and directional awareness, allowing the navigation policy to maintain task-relevant spatial context internally. Extensive evaluations in both simulated and real-world environments demonstrate that our approach empowers the robot to robustly traverse clutters and escape maze dead-ends without the reliance on continuous external goal references, validating the effectiveness of the learned spatial awareness alongside zero-shot sim-to-real transfer ability.  

However, we observe that GUIDE occasionally misses traversable openings in confined spaces, potentially due to the limited field of view of the onboard depth sensor. Future work will explore active perception and additional sensor modalities to expand the robot's local spatial observations. Additionally, we will investigate how to extend GUIDE to larger-scale navigation scenarios while maintaining reliable internal spatial awareness.

\bibliography{references}  %
\clearpage
\appendices

\end{document}